\documentclass[letterpaper]{article}
\usepackage{aaai2026}
\usepackage{times}
\usepackage{helvet}
\usepackage{courier}
\usepackage[hyphens]{url}
\usepackage{graphicx}
\usepackage{natbib}
\usepackage{caption}
\usepackage{algorithm}
\usepackage{algorithmic}
\usepackage{amsmath,amssymb}
\usepackage{newfloat}
\usepackage{listings}

\DeclareCaptionStyle{ruled}{labelfont=normalfont,labelsep=colon,strut=off}
\floatstyle{ruled}
\newfloat{listing}{tb}{lst}{}
\floatname{listing}{Listing}

\title{Project Aletheia: Verifier-Guided Distillation of Backtracking for Small Language Models}

\author{
Aradhya Dixit\textsuperscript{1}\thanks{These authors contributed equally.}, 
Tianxi Liang\textsuperscript{2},
Jai Telang\textsuperscript{3} \\
\textsuperscript{1}Wake Technical Community College,
\textsuperscript{2}Cornell University,
\textsuperscript{3}Algoverse \\
adixit1@my.waketech.edu, tianxi730@gmail.com, jptelang17@gmail.com
}

\begin{document}
\maketitle

\begin{abstract}
Small Language Models (SLMs, under 10B parameters) are attractive for private, on-device deployment, yet they frequently fail on strict constraint-satisfaction problems due to linear, overconfident reasoning traces that do not recover from early mistakes. We introduce \emph{Verifier-Guided Distillation}, a training protocol that transfers the \emph{process of error repair}---explicit conflict detection and backtracking---rather than only correct final answers. By training a 7B model on verified reasoning traces that include mistakes and self-corrections, we show that latent verification behavior can emerge in small models, enabling them to occasionally stop, detect contradictions, and revise earlier assumptions.

\end{abstract}

\section{Introduction}
Small Language Models (SLMs, <10B parameters) are attractive for on-device deployment but often fail on strict constraint-satisfaction tasks. For Boolean satisfiability (SAT), early assignment errors propagate, producing invalid solutions, since even a single falsified clause invalidates the entire assignment \citep{davis1962machine}.

Frontier models achieve strong reasoning performance by producing intermediate hypotheses and iteratively revising them before committing to a final answer \citep{openai_o1_system_card,google_gemini25}, akin to ``slow thinking'' in cognitive science \citep{kahneman2011thinking}. Standard supervised training, however, emphasizes monotonic solution paths, encouraging overconfident continuation even after errors \citep{wei2022cot,kojima2022zeroshot,wang2022selfconsistency}.

We argue that SLM failures are not just capacity-limited but also reflect a \emph{training-data bias}: most corpora overrepresent flawless reasoning and underrepresent recoverable errors. We propose \textbf{Project Aletheia}, which distills the \emph{process of error repair}—detecting conflicts, revising state, and continuing—into small models. By supervising SLMs with traces including both mistakes and corrective backtracking, we show latent verification behavior can emerge even under resource constraints.

Our contributions are: (i) introducing \emph{Verifier-Guided Distillation} for transferring backtracking and self-correction into SLMs; (ii) constructing a verified SAT training dataset with explicit conflict and repair steps, alongside a linearized control dataset; and (iii) demonstrating that even a single-epoch LoRA fine-tune can induce observable backtracking behavior in previously monotonic models.

\section{Method: Verifier-Guided Distillation}
We aim to distill a \emph{procedure}---conflict detection and state revision---into a small model, rather than distilling only correct final assignments.
Our pipeline has three stages: (i) synthesizing backtracking-rich teacher traces, (ii) symbolic verification of trace correctness, and (iii) controlled distillation under a matched control/treatment design.

\subsection{Problem Setup (SAT as Constraint Satisfaction)}
We consider Boolean satisfiability (SAT) instances given in conjunctive normal form (CNF):
\begin{equation}
F \;=\; \bigwedge_{i=1}^{m} C_i,
\end{equation}
where each clause $C_i$ is a disjunction of literals,
$C_i = \bigvee_{j=1}^{k_i} \ell_{ij}$, and each literal $\ell_{ij}$ is either a variable $x_p$ or its negation $\neg x_p$.
A (partial) assignment is a mapping $a: \{x_1,\dots,x_n\}\rightarrow\{0,1,\bot\}$ where $\bot$ denotes unassigned.
A \emph{conflict} occurs when some clause is falsified under the current assignment:
\begin{equation}
\exists i \;\; \text{s.t.}\;\; C_i(a_t) = 0,
\end{equation}
i.e., every literal in $C_i$ evaluates to false under $a_t$ (with no remaining unassigned literal that could make it true).
We represent solver state as $s_t = (a_t, H_t)$, where $a_t$ is the current (partial) assignment at step $t$ and $H_t$ is a trace of emitted reasoning steps.\subsection{Verification and Dataset Construction}
Teacher traces may hallucinate intermediate steps; we enforce correctness only on final assignments using PySAT \citep{ignatiev2018pysat}. From 500 generation attempts, 496 traces (99.2\%) passed verification.  
\textbf{Caveat:} The dataset is small and may not cover all types of conflicts (single- vs multi-variable, early vs delayed), and traces reflect the heuristics of the teacher model. These factors may limit generalization but are sufficient for initial proof-of-concept distillation.

\subsection{Negative Constraint Injection (Backtracking-Supervision Trace Synthesis)}
To generate supervision that explicitly encodes self-correction, we prompt Gemini 2.5 Pro \citep{google_gemini25} to act as a backtracking engine rather than a direct-answer oracle.
At each decision step, the teacher proposes an assignment, runs an explicit constraint check, and, upon detecting a conflict, emits a standardized token \texttt{[CONFLICT]} followed by a revision routine.

\paragraph{Teacher policy.}
We treat the teacher trace as a sequence of actions $\pi_T$ producing tokens and assignment operations.
At step $t$, the teacher chooses either:
\begin{itemize}
\item \textbf{Decide:} select an unassigned variable $x_p$ and set $a_t(x_p)\leftarrow v$ for $v\in\{0,1\}$.
\item \textbf{Verify:} evaluate all clauses under $a_t$; if any clause is falsified, emit \texttt{[CONFLICT]} and identify a violated clause.
\item \textbf{Backtrack:} revert some prior decision(s), e.g. undo the last assignment or a short window of recent assignments:
\begin{equation}
a_{t+1} \;=\; \text{Revert}(a_t; \, \Delta),
\end{equation}
where $\Delta$ encodes which variable(s) to unassign (or flip and then re-check).
\end{itemize}

\paragraph{Why ``negative'' constraints help.}
Standard chain-of-thought distillation overrepresents \emph{successful} partial assignments, providing weak supervision for what to do when a clause becomes falsified \citep{wei2022cot}.
Negative Constraint Injection forces the teacher to surface failure states and repair moves, creating training signals aligned with the desired behavior:
\begin{equation}
\text{If}\;\exists i:\, C_i(a_t)=0 \;\Rightarrow\; \text{emit \texttt{[CONFLICT]} and revise } a_t.
\end{equation}

\paragraph{Trace format.}
Each teacher trace is serialized as a structured natural-language log containing:
(i) the current assignment state,
(ii) clause-check outcomes,
(iii) an explicit conflict marker,
and (iv) corrective backtracking steps.
This yields trajectories that encode both incorrect hypotheses and their resolution.

\subsection{Symbolic Verification and ``Golden'' Dataset Construction}
Because teacher traces can hallucinate satisfying assignments or misreport clause evaluations, we enforce correctness with a symbolic judge.
For each SAT instance $F$ and final assignment $a^{\star}$ produced by a teacher trace, we verify:
\begin{equation}
\forall i\in\{1,\dots,m\}, \;\; C_i(a^{\star}) = 1.
\end{equation}
We implement this judge using the PySAT toolkit \citep{ignatiev2018pysat}.
Only traces whose final assignments satisfy the CNF under exact evaluation are retained.
From 500 generation attempts, 496 pass verification (99.2\%), forming our ``golden'' dataset.

\begin{algorithm}[t]
\caption{Symbolic verification of a teacher trace (PySAT).}
\label{alg:verify}
\begin{algorithmic}[1]
\REQUIRE CNF formula $F$; teacher-proposed final assignment $a^\star$; PySAT solver
\ENSURE $\texttt{verified}\in\{0,1\}$
\STATE Encode $F$ as PySAT CNF object
\STATE Translate $a^\star$ into a list of signed literals (e.g., $x=1 \mapsto +x$, $x=0 \mapsto -x$)
\STATE $\texttt{verified} \leftarrow \texttt{Solve}(F \wedge a^\star)$
\RETURN $\texttt{verified}$
\end{algorithmic}
\end{algorithm}

\subsection{Control vs.\ Treatment: Matched Ablation of Backtracking Signal}
To isolate whether backtracking supervision (not just exposure to SAT solutions) induces verification loops, we derive two training corpora from the same verified traces.

\paragraph{Treatment (Aletheia traces).}
We retain the full sequence including incorrect proposals, clause-check commentary, explicit \texttt{[CONFLICT]} markers, and corrective backtracking.

\paragraph{Control (Linearized traces).}
We remove conflict/backtracking spans and stitch the remaining steps into a monotonic reasoning path that ends in the same verified satisfying assignment.
If a trace contains:
\[
\cdots \rightarrow \texttt{[CONFLICT]} \rightarrow \text{Backtrack} \rightarrow \text{Revised proposal} \rightarrow \cdots
\]
we drop the \texttt{[CONFLICT]} block and any reverted steps, keeping the revised proposal and subsequent successful steps.
Thus, control and treatment share (i) the same task distribution, (ii) the same final answers, and (iii) similar token budgets, differing primarily in whether the student sees explicit self-correction patterns.

\begin{listing}[tb]
\caption{Pseudo-code for trace linearization (control construction).}
\label{lst:linearize}
\begin{lstlisting}
def linearize(trace_lines):
    out = []
    skip = False
    for line in trace_lines:
        if "[CONFLICT]" in line:
            skip = True
            continue
        if skip:
            # skip backtracking narration + reverted steps
            if "New Proposal" in line or "Revised" in line:
                skip = False
                out.append(line)
            continue
        out.append(line)
    return out
\end{lstlisting}
\end{listing}

\subsection{Student Training (LoRA Distillation Under Edge Constraints)}
We fine-tune Qwen-2.5-7B-Instruct \citep{qwen25_7b_instruct} via supervised imitation of teacher traces, using parameter-efficient LoRA adapters \citep{hu2021lora} and 4-bit quantization to approximate edge deployment constraints \citep{dettmers2023qlora,huggingface_4bit_blog}.

\paragraph{Objective.}
Let $y = (y_1,\dots,y_T)$ be the target trace tokens and $x$ be the SAT instance prompt.
We minimize autoregressive cross-entropy:
\begin{equation}
\mathcal{L}(\theta) \;=\; -\sum_{t=1}^{T} \log p_{\theta}(y_t \mid x, y_{<t}).
\end{equation}
In the treatment condition, $y$ includes explicit \texttt{[CONFLICT]} and backtracking tokens; in the control condition, these tokens are absent.

\paragraph{LoRA parameterization.}
For a linear projection $W\in\mathbb{R}^{d\times k}$, LoRA learns low-rank updates:
\begin{equation}
W' \;=\; W + \Delta W,\quad \Delta W = BA,\quad
A\in\mathbb{R}^{r\times k},\; B\in\mathbb{R}^{d\times r},
\end{equation}
with rank $r=16$ and scaling $\alpha=16$ \citep{hu2021lora}.
We apply LoRA to all major projection layers ($Q,K,V,O,\text{Gate},\text{Up},\text{Down}$).

\paragraph{Compute and tooling.}
Training is performed on a single NVIDIA A100-SXM4-40GB GPU using Unsloth \citep{unsloth_github}.
Control and treatment adapters are trained with identical hyperparameters and training budgets.

\paragraph{Hyperparameters.}
Learning rate $2\times 10^{-4}$, epochs=1, LoRA rank $r=16$, LoRA scaling $\alpha=16$.

\begin{figure}[t]
\centering
\includegraphics[width=0.95\columnwidth]{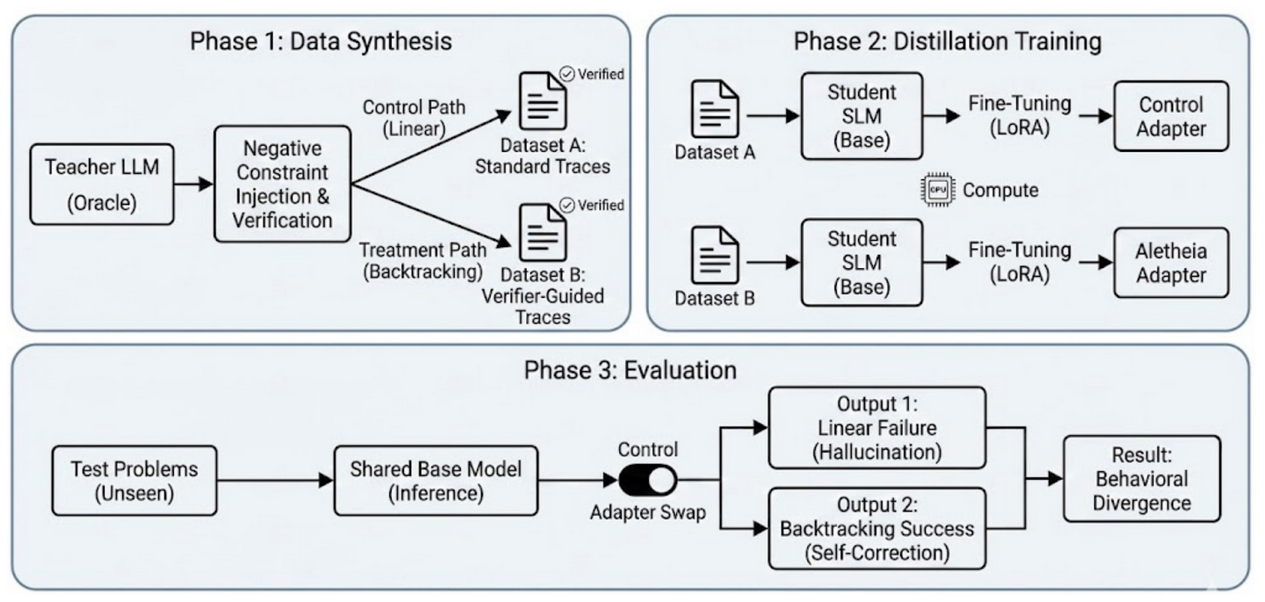}
\caption{End-to-end Verifier-Guided Distillation pipeline.
Teacher generates backtracking traces under Negative Constraint Injection; PySAT verifies final assignments; control vs.\ treatment adapters are trained and evaluated via adapter hot-swapping.}
\label{fig:pipeline}
\end{figure}

\section{Evaluation}
We evaluate on 40 held-out SATBench problems \citep{wei2025satbench} using a shared base model and adapter hot-swapping to compare control vs.\ treatment under identical decoding.
We use temperature $0.6$ to encourage exploration, making verification behavior observable rather than suppressed by greedy decoding \citep{holtzman2019curious}.

\subsection{Backtracking Event Rate (Primary Behavioral Metric)}
Our primary metric measures whether the model triggers an explicit verification loop.
Define an event indicator for a generated response $R$:
\begin{equation}
\mathbf{1}_{\text{bt}}(R) \;=\;
\begin{cases}
1, & \text{if } R \text{ contains \texttt{[CONFLICT]}}\\
   & \text{and an explicit revision step},\\
0, & \text{otherwise.}
\end{cases}
\end{equation}
We report the backtracking event rate over $N=40$ instances:
\begin{equation}
\text{BER} \;=\; \frac{1}{N}\sum_{i=1}^{N} \mathbf{1}_{\text{bt}}(R_i).
\end{equation}
We operationalize ``explicit revision'' as (i) a backtracking phrase (e.g., ``Backtracking'', ``Reverting state'') and (ii) a changed assignment proposal in subsequent lines.

\subsection{Correctness (Secondary Metric)}
Separately from behavior, we optionally verify the final assignment produced by the model with the same PySAT judge \citep{ignatiev2018pysat}, yielding a correctness indicator $\mathbf{1}_{\text{sat}}(a^\star)$.
In this 4-page version, we emphasize BER as the key evidence of distilled self-correction.

\section{Results}
We evaluate two adapters trained on identical base models and data distributions:
a \emph{Control} adapter trained on linearized traces, and an \emph{Aletheia} adapter trained on full backtracking traces.
Both are tested under identical decoding settings on 40 held-out SATBench instances, using adapter hot-swapping to isolate the effect of supervision.

\subsection{Backtracking Event Rate}
The primary metric is BER: the fraction of inference traces that contain (i) \texttt{[CONFLICT]} and (ii) a subsequent state revision step.

The Control model exhibits no observable verification behavior.
Across all 40 test instances, it generates monotonic traces without emitting conflict markers or revising earlier assignments:
\[
\text{BER}_{\text{control}} = \frac{0}{40} = 0.
\]

In contrast, the Aletheia-trained model triggers explicit verification behavior in 2 out of 40 cases:
\[
\text{BER}_{\text{aletheia}} = \frac{2}{40} = 0.05.
\]
In these instances, the model emits a \texttt{[CONFLICT]} signal, retracts a prior assignment, and proposes an alternative value.
Since the two adapters differ only in exposure to failure-and-repair supervision, this behavior is attributable to the distillation signal.

\subsection{Qualitative Trace Analysis}
Figure~\ref{lst:killshot} incorporates an excerpted qualitative trace (same structure as Listing~\ref{lst:killshot}), showing (i) a hypothesis proposal, (ii) explicit conflict detection, and (iii) localized revision via backtracking.

\begin{listing}[tb]
\caption{Representative Aletheia inference trace exhibiting explicit conflict detection and backtracking.}
\label{lst:killshot}
\begin{lstlisting}
1. Initial State:
   All variables unassigned.
   Assignment set = {}.

2. Propose Assignment:
   Based on Constraint 1 (x3 = false),
   assign x3 = False.
   -> Current State: {x3: False}

3. Check Assignment:
   Verifying against constraints...
   - Constraint 1 satisfied.
   - Constraint 2 violated.

   [CONFLICT] Constraint violation detected.
   Backtracking...

   Reverting assignment x3 = False.

4. New Proposal:
   Assign x3 = True.
   -> Current State: {x3: True}
   ...
\end{lstlisting}
\end{listing}

\subsection{Summary of Empirical Findings}
Empirically, we observe a clear behavioral contrast:
linear supervision yields zero backtracking behavior, while failure-aware supervision yields non-zero, structured verification behavior.
Although the absolute frequency is low, the appearance of any explicit self-correction after a single-epoch LoRA fine-tune supports the claim that the \emph{mechanism} of backtracking can be transferred through supervised distillation.

\section{Discussion and Future Work}
\subsection{Interpretation}
These results support a data-topology hypothesis: SLM reasoning failures are partly due to limited exposure to \emph{recoverable errors} during training.
Conventional reasoning supervision can implicitly teach that ``reasoning is monotonic'' \citep{wei2022cot}, whereas Aletheia supervision teaches a conditional control rule: \emph{when a contradiction is detected, pause and revise state}.
The exclusive appearance of conflict-and-repair behavior in the Aletheia condition suggests that self-correction is a learnable routine, not only an emergent artifact of scale or test-time compute \citep{openai_o1_system_card,google_gemini25}.

\subsection{Limitations}
This study demonstrates feasibility rather than robustness. First, the Backtracking Event Rate (BER) is low (5\%), so self-correction is rare. Second, the evaluation is limited to SAT; generalization to other domains is untested. Third, supervised imitation does not directly incentivize the model to choose backtracking when next-token likelihood is low. Fourth, the current behavioral metric is string-based; parser-based detection and end-to-end satisfiability checks would yield richer evaluation.

\textbf{Data limitations:} Our training dataset is small (496 verified traces) and reflects the heuristics of the teacher model, which may bias backtracking patterns. Conflict types are not systematically controlled, and only final assignments are verified—intermediate steps may contain minor inconsistencies. Control traces are linearized from treatment traces, sharing final assignments, which could affect comparisons. These factors constrain generalization but suffice for a proof-of-concept demonstration.

\subsection{Future Work}
\paragraph{RL over repairs (from syntax to strategy).}
Once the model has learned the \emph{syntax} of backtracking via distillation, reinforcement learning can optimize its \emph{usage}.
A simple reward can be issued when a backtracking step resolves a contradiction and yields a satisfying assignment (verified by PySAT), turning rare self-correction into a deliberate strategy \citep{ignatiev2018pysat}.
This mirrors a broader shift toward training models to reason with verifiable intermediate checks rather than purely imitating fluent traces \citep{wang2022selfconsistency}.

\paragraph{Curriculum over failure modes (deeper search).}
We can progressively introduce (i) delayed conflicts (contradictions only visible after multiple decisions), (ii) multi-variable rollbacks, and (iii) adversarial ``tempting'' assignments that appear locally consistent but fail globally.
Curricula of increasing repair depth would push beyond one-step revisions toward systematic search-like behavior, more aligned with classical SAT procedures \citep{davis1962machine}.

\paragraph{Stronger behavioral instrumentation.}
Replace string heuristics with a finite-state detector that tracks explicit assignment diffs (e.g., extracting $\{x_i\!\mapsto\!v\}$ from text) and counts (a) retractions, (b) flips, and (c) verified satisfiable completions.
This would allow reporting richer metrics such as ``conflict precision'' (fraction of emitted conflicts that correspond to an actually falsified clause under the model-stated assignment).

\paragraph{Generalization beyond SAT.}
Verifier-guided distillation is domain-agnostic: any task with an external judge can provide the same ``detect $\rightarrow$ repair'' supervision.
Natural next targets include arithmetic proof checking, program synthesis (unit tests as a verifier), and symbolic planning (state transition checkers), where backtracking is also fundamental.

\paragraph{Edge deployment implications.}
Because the student runs under strict parameter and quantization constraints, this approach offers a complementary path to logic-capable, privacy-preserving agents without proprietary inference-time scaling \citep{hu2021lora,dettmers2023qlora}.
Distilling \emph{how to recover} (not just what to answer) may be a practical route to robust on-device reasoning.

\bibliography{aaai2026}
\end{document}